\documentclass[journal]{IEEEtran}

\usepackage{amsmath}    % For \eqref and other math environments
\usepackage{amssymb}    % For symbols
\usepackage{amsfonts}   % For math fonts
\usepackage{bm}         % For bold math

\usepackage{CJKutf8}

\usepackage{graphicx}   % For including graphics
\usepackage{epstopdf}   % For EPS to PDF
\usepackage{float}      % For [H] specifier
\usepackage{tikz}       % For drawing diagrams
\usetikzlibrary{positioning, fit, calc, shapes.geometric, arrows.meta}

\usepackage[caption=false,font=footnotesize]{subfig}   % Subfigures (no caption package needed)

\usepackage{booktabs}   % For professional-looking tables
\usepackage{multirow}   % For table cells spanning multiple rows

\usepackage{times}      % Times font
\usepackage{indentfirst}% Indent first paragraph
\usepackage{color}      % Colored text

\usepackage{cite}       % For compact citations
\usepackage{enumerate}  % For custom numbered lists

\usepackage[linesnumbered,ruled,vlined]{algorithm2e}  % Best for algorithm boxes

\usepackage{amsthm}
\newtheorem{definition}{Definition}

\usepackage[
    colorlinks=false,       % Keep text black
    pdfborder={0 0 1},      % Draw a small box around links
    citebordercolor={0 1 0},% Green box for citations
    linkbordercolor={1 0 0},% Red box for refs (figures, tables, equations)
    urlbordercolor={0 0 1}  % Blue box for URLs
]{hyperref}

\begin{document}

\title{Importance-aware Topic Modeling for Discovering Public Transit Risk from Noisy Social Media}

\author{%
  Fatima Ashraf,
    Muhammad Ayub Sabir, Jiaxin Deng, Junbiao Pang\textsuperscript{$*$}, Haitao Yu%
  \thanks{\textsuperscript{*}Corresponding author. E-mail: \href{mailto:junbiao_pang@bjut.edu.cn}{junbiao\_pang@bjut.edu.cn}}%
  \thanks{Fatima, Sabir, Deng, and J.Pang are with the Faculty of Information Technology, Beijing University of Technology, Beijing 100124, China.\\
  Haitao Yu is with Beijing Intelligent Transportation Development Center, Beijing 100161, China}%

}

\maketitle

\begin{abstract}
Urban transit agencies increasingly turn to social media to monitor emerging service risks such as crowding, delays, and safety incidents, yet the signals of concern are sparse, short, and easily drowned by routine chatter. We address this challenge by jointly modeling linguistic interactions and user influence. First, we construct an influence-weighted keyword co-occurrence graph from cleaned posts so that socially impactful posts contributes proportionally to the underlying evidence. The core of our framework is a Poisson Deconvolution Factorization (PDF) that decomposes this graph into a low-rank topical structure and topic-localized residual interactions, producing an interpretable topic--keyword basis together with topic importance scores. A decorrelation regularizer \emph{promotes} distinct topics, and a lightweight optimization procedure ensures stable convergence under nonnegativity and normalization constraints. Finally, the number of topics is selected through a coherence-driven sweep that evaluates the quality and distinctness of the learned topics. On large-scale social streams, the proposed model achieves state-of-the-art topic coherence and strong diversity compared with leading baselines. The code and dataset are publicly available at~\url{https://github.com/pangjunbiao/Topic-Modeling_ITS.git}

\end{abstract}

\begin{IEEEkeywords}
Intelligent Transportation Systems, Social Media Mining, Weak Signal Detection, Topic Modeling, Graph Factorization, Urban Transit Analytics
\end{IEEEkeywords}

% origianl one :
\section{Introduction}

Urban public transportation is a critical public service whose performance is shaped not only by physical disruptions but also by rider's day-to-day experiences \cite{li2019similarusermining, pang2024finding, qi2019urban}. While operational systems capture structured signals (e.g., GPS traces, sensor telemetry, incident logs), a large volume of weak yet actionable feedback appears in social media streams such as Weibo, where riders report delays, comfort, payment issues, and safety concerns in real time \cite{zuo2018extraction,tse2018social, pang2016robust}. Mining such user-generated content can improve situational awareness and support proactive service management.

Social-media posts are sparse, short, noisy, and semantically fragmented, which challenges standard topic models and monitoring pipelines \cite{pang2018two,liu2018topic, laureate2023systematic}. Classical probabilistic approaches (e.g., LDA and its extensions) capture coarse themes but ignore user influence and often yield redundant or diffuse topics. For Weibo specifically, neural and heuristic systems—including binary relevance detection \cite{chen2018detecting}, lexicon-based extraction and visualization of negative traffic events \cite{zuo2018extraction}, and fusion with news via word-embedding alignment \cite{lu2018sensing, sun2020identifying}—have advanced topic detection, yet they typically require labeled data, depend on fixed schemas, or lack interpretability for operations. Consequently, existing methods still struggle to (i) separate structured topical signal from background interactions and (ii) prioritize socially \emph{influential} content rather than merely frequent content.

Building on these observations, we argue that effective weak-signal mining should integrate three complementary elements: (i) a \emph{keyword interaction view} that retains local co-occurrence structure from short texts; (ii) an explicit \emph{influence model} so impactful posts contribute more than incidental chatter; and (iii) an \emph{interpretable factorization} that separates low-rank topical structure from residual interactions while discouraging topic overlap. This perspective motivates the development of a unified framework featuring a graph-based formulation with an influence-weighted co-occurrence matrix and a Poisson Deconvolution Factorization (PDF) objective specifically tailored for sparse, nonnegative data.

\begin{figure*}[!t]
    \centering
    \includegraphics[width=1.0\textwidth]{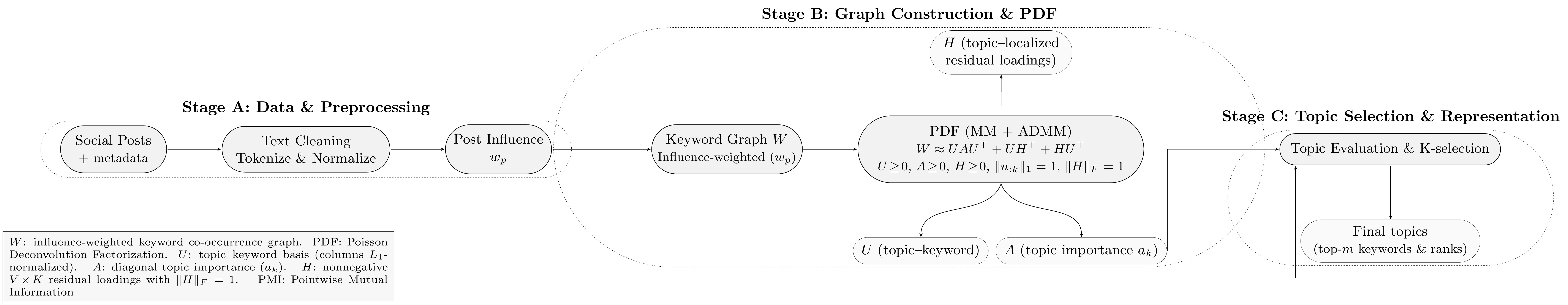}
    \caption{Overview of the proposed influence-weighted topic modeling framework}
    \label{fig:framework}
\end{figure*}

Guided by the above perspective, the proposed framework operates in three main stages, as illustrated in Fig.~\ref{fig:framework}.
In \emph{Data and Preprocessing}, social posts and their metadata are cleaned, tokenized, and normalized to remove noise while preserving meaningful lexical patterns.
Each post is assigned an \emph{influence weight} derived from engagement and reach statistics, ensuring that user feedback with greater operational impact contributes more strongly to subsequent analysis. In \emph{Graph Construction and Poisson Deconvolution Factorization}, an influence-weighted keyword co-occurrence graph is constructed from \emph{unordered adjacent bigrams} aggregated across posts. 

This graph is then decomposed through Poisson Deconvolution Factorization, which separates a low-rank topical structure from sparse, topic-localized residual interactions, yielding an interpretable topic--keyword basis and topic importance scores. A decorrelation regularizer promotes distinct topics, while an efficient optimization strategy combines multiplicative updates for the main factors with an ADMM routine for the residual component under nonnegativity and normalization constraints. Finally, in topic selection and representation, the number of topics is determined through a coherence-driven sweep over the social media corpus, evaluating both the quality and distinctness of the learned topics.

The main contributions of this work are summarized as follows:

\begin{itemize}
    % \item \textbf{Influence-weighted keyword graph.} 
    \item We construct an influence-weighted co-occurrence graph that integrates user engagement and reach, allowing socially impactful posts to contribute proportionally to the semantic structure.

    % \item \textbf{Poisson Deconvolution Factorization.} 
    \item We formulate a Poisson-based factorization model that separates the influence-weighted graph into a low-rank topical structure and sparse residual interactions under nonnegativity and decorrelation constraints.

    % \item \textbf{Efficient optimization.} 
    \item We design a lightweight optimization scheme combining multiplicative updates for the main factors with an ADMM routine for the constrained residual component, ensuring stability and scalability.

    % \item \textbf{Topic selection and interpretability.} 
    \item We provide an interpretable topic-selection and representation procedure that selects an appropriate topic number via a coherence-driven sweep over posts and ranks topics by their learned importance
\end{itemize}

\section{Related Work} \label{sec:RW}

\subsection{Topic and Temporal Modeling for Noisy Social Streams}
Short social posts pose well-known challenges—sparsity, informal vocabulary, and rapid drift—which degrade classic bag-of-words topic models \cite{murshed2023short, garcia2025concept}. Foundational probabilistic topic models such as LDA~\cite{blei2003latent} and HDP~\cite{teh2006hierarchical} remain central for mining themes in social media \cite{blei2003latent,teh2006hierarchical,2017Probabilistic}, and neural probabilistic variants such as GSM~\cite{miao2017discovering} extend this line. Recent surveys show that neural topic models (NTMs) and contextualized representations now dominate short-text analysis, enhancing coherence and robustness on social media corpora \cite{wu2024survey,zhao2021topic}. Yet the noisy and rapidly evolving nature of microblogs still limits their effectiveness. To address temporal and sparsity issues, subsequent work introduces dynamic and supervised models that couple text with time series \cite{park2015supervised}, and sparsity-aware regularization to improve coherence \cite{he2017efficient}. In parallel, high-dimensional and network-based approaches aim to detect emerging “hot topics’’ from weak signals \cite{chen2020highdimensional}, while cross-modal studies emphasize resilience to weak labels and drift \cite{lee2018cleannet,salakhutdinov2011learning}. Despite these advances, many approaches still overlook user influence and do not explicitly separate topical signal from residual structure—gaps we address with an influence-weighted graph and PDF.

\subsection{Structural Signals: Users, Influence, and Graph Views}
Beyond bag-of-words text, \emph{structural signals}—who speaks, how posts connect, and how words co-occur—are critical for robust topic discovery in noisy social streams \cite{panchendrarajan2023topic}. Global text graphs that link documents and words propagate information over co-occurrence structure and improve performance on sparse texts (e.g., TextGCN) \cite{yao2019graph}. Building on this paradigm, graph topic models learn topics directly from document/word relations via neural message passing, yielding more coherent, separable topics under short-text sparsity \cite{zhou2020gtm,shen2021docgraph}. Complementary network-aware strategies surface emerging “hot topics’’ from thin signals by leveraging interaction and co-usage structure \cite{chen2020highdimensional}. On the \emph{user side}, influence and authorship structure modulate topical salience: similar-user mining via mutual attention/fan overlap enhances topical similarity \cite{li2019similarusermining}, while graph author–topic models propagate topic signals along author/venue relations to stabilize estimates on sparse corpora \cite{zhang2022vgatm}. Heterogeneous or multi-view text graphs that combine word–word and document–word edges (and optionally metadata) provide additional robustness to noise and drift \cite{linmei2019heterogeneous,zhang2022heterogeneous}. These findings motivate our design: an \emph{influence-weighted keyword co-occurrence graph} that captures word–word structure while weighting evidence by post-level impact, paired with a \emph{factorization over that graph} to separate structured topical signal from residual interactions in an interpretable manner.

\subsection{Social Media for Transit Operations}
A parallel line of work leverages social streams to augment transit monitoring and operations. Early studies established that tweets can support real-time traffic/incident detection with timeliness and coverage complementary to conventional sensors \cite{gu2016twitter}. Subsequent systems mine Twitter to monitor traffic and road conditions at scale, demonstrating practical pipelines for extraction and visualization \cite{putra2022traffic}. For transit specifically, recent analyses use social media to estimate performance metrics and compare rider-reported issues with agency-reported indicators \cite{kamga2023utilizing}, and to uncover complaint categories and service grievances across cities \cite{pullanikkat2024utilizing}. Topic modeling on Twitter has been used to track shifts in public opinion toward transit during the COVID-19 period \cite{taleqani2021using}, while broader work shows social data can function as passive geo-participation, aligning with survey signals in transportation contexts \cite{lock2020social}. Studies on Weibo similarly reveal public concerns and sentiment around transport and mobility topics \cite{yang2022topic,zha2023social}. These findings motivate and treating social media posts as a complementary sensing layer for for modern transit operations. 

\begin{table*}[t]
\centering
\caption{Statistics of the collected social media datasets by city.}
\scriptsize % Further reduced the font size
\renewcommand{\arraystretch}{1.2} % Increased row height for better spacing
\begin{tabular}{l p{1.6cm} p{1.6cm} p{1.6cm} p{1.6cm} p{1.6cm} p{1.6cm} p{1.7cm} p{1.7cm}} % Increased the width of the last two columns slightly
\toprule
& \multicolumn{1}{c}{Posts (N)} & \multicolumn{1}{c}{Unique accounts} & \multicolumn{1}{c}{Posts/account} & \multicolumn{1}{c}{Max posts} & \multicolumn{1}{c}{Likes} & \multicolumn{1}{c}{Comments} & \multicolumn{1}{c}{Followers} & \multicolumn{1}{c}{Duplicates} \\
City & & & \multicolumn{1}{c}{median [IQR]} & \multicolumn{1}{c}{by an acct.} & \multicolumn{1}{c}{median [IQR]} & \multicolumn{1}{c}{median [IQR]} & \multicolumn{1}{c}{median [IQR]} & \multicolumn{1}{c}{(n, \%)} \\
\midrule
Beijing & 812 & 664 & 1 [1--1] & 38 & 1 [0--3] & 1 [0--3] & 304 [86--8340]  & 7, 0.9\% \\
Shanghai & 588 & 518 & 1 [1--1] & 13 & 1 [0--3] & 0 [0--2] & 284 [117--2684] & 8, 1.4\% \\
Xiamen & 673 & 444 & 1 [1--1] & 30 & 2 [0--5] & 1 [0--4] & 434 [159--61344] & 2, 0.3\% \\
\bottomrule
\end{tabular}
\label{tab:data-stats}
\end{table*}

\section{Methodology}\label{sec:method}

Social media posts differ from traditional documents: they are short, noisy, and heterogeneous in both content and user influence. Individual posts often carry weak linguistic evidence (few tokens, context dependence), while hyperlinks, mentions, emojis, and slang introduce lexical noise. At the same time, platform dynamics can transiently elevate some users (e.g., bursts of engagement), making post impact highly uneven. These properties motivate deliberate preprocessing and an influence-aware co-occurrence representation used by our topic model.

\begin{definition}[Post]\label{def:post}
Let the set of posts be $\mathcal{P}=\{1,2,\ldots,N\}$. For each $p\in\mathcal{P}$, let $(u_p, x_p, m_p)$ denote the author identifier $u_p$, the raw text content $x_p$, and optional metadata $m_p$ such as the author’s follower count $F_p$ and post-level interaction statistics (e.g., counts and timestamps of reposts, likes, and comments).
\end{definition}

Given the characteristics above, three primary challenges arise in modeling such data.

\begin{itemize}
  \item \raggedright \textbf{Weak textual signal.} Short, context-dependent posts yield sparse evidence; we therefore normalize text and merge frequent multi-word expressions to densify the semantic signal.
  \item \raggedright \textbf{High lexical noise.} Informal variants, emojis, URLs, and user handles can dominate counts; we apply robust cleaning with domain-term protection to preserve meaningful tokens while filtering artifacts.
  \item \raggedright \textbf{Role fluidity.} User influence fluctuates sharply across posts; we incorporate engagement- and reach-based weights so each post’s contribution reflects its actual impact on the discourse.
\end{itemize}

\subsection{Dataset}\label{sec:data}
We curate three city–specific corpora from Chinese social media (Weibo) posts referring to public transport in \textbf{Beijing}, \textbf{Shanghai}, and \textbf{Xiamen}. Posts were collected from a mix of official transit accounts and individual users via publicly accessible web interfaces, in accordance with the platform’s terms of service.

Each file shares an identical schema: \textbf{post\_id} — account identifier of the authoring account (used only for aggregation; not released); \textbf{timestamp} — post creation time as provided by the platform; \textbf{text} — full post body, including hashtags (URLs/mentions/emojis are preserved in the raw data); \textbf{likes} — number of likes at crawl time; \textbf{comments} — number of comments at crawl time; \textbf{followers} — follower count of the authoring account at crawl time; \textbf{platform} — platform label inferred from the source.

In Table ~\ref{tab:data-stats}, For each city, we report: total posts; number of \emph{unique accounts} (distinct \texttt{post\_id}); \emph{Posts/account} summarized by \emph{median~[IQR]}, where the median is the 50th percentile across accounts and IQR is the interquartile range (75th--25th percentile); the \emph{Max posts by an account}; and engagement summaries (likes, comments, followers) again as \emph{median~[IQR]} across posts. We additionally report the number and share of removed duplicates.

Other public Chinese social media corpora used in transport studies typically (i) aggregate across multiple cities, (ii) sample on global keywords (e.g., “bus”, “subway”) rather than city–specific entities, (iii) omit account-level fields (follower counts) needed for influence weighting, and (iv) do not preserve narrow, city-scoped time windows. In contrast, our dataset is \emph{city-resolved}, \emph{time-bounded}, and \emph{schema-complete} for our modeling pipeline (text, engagement, and account reach), enabling apples-to-apples comparisons across cities and supporting our influence-aware co-occurrence modeling.

\subsection{Data Preprocessing}\label{sec:preprocess}

Social media posts include emojis, hashtags, user handles, and links that can obscure lexical regularities. Text is standardized and normalized prior to vocabulary construction as follows:
\begin{itemize}
    \item \raggedright control symbols are removed; punctuation and case are normalized; emojis and special characters (e.g., ``@'', ``\#'', ``\&'', ``*'') are stripped unless retained as domain tokens;
    \item \raggedright user handles and URLs are removed; hashtags are mapped to their lexical content (e.g., ``\#service\_delay'' $\rightarrow$ ``service delay'') when informative and otherwise dropped;
    \item \raggedright language-appropriate tokenization is applied (e.g., word segmentation for Chinese), and common variants, abbreviations, and slang are normalized to a canonical form (e.g., ``APP'' $\leftrightarrow$ ``app''; \begin{CJK}{UTF8}{gbsn}“蓝色香菇”\end{CJK} $\rightarrow$ \begin{CJK}{UTF8}{gbsn}“难受想哭”\end{CJK} when they denote the same concept).
\end{itemize}
 
The output is, for each \(p \in \mathcal{P}\), a cleaned token sequence \(s_p = (w_{p,1}, \dots, w_{p,L_p})\) and a deduplicated vocabulary \(\mathcal{V}\). These sequences are the sole textual inputs used downstream to construct the static keyword graph \(W\).

\subsection{User Weight Construction}\label{sec:UWC}

Each post is assigned a static influence weight $w_p$ so that socially impactful content contributes more to downstream topic modeling. The construction combines (i) engagement normalized by audience reach and (ii) an arrival–rate proxy derived from inter–comment gaps.

Let $c_p,l_p,r_p$ denote the counts of comments, likes, and reposts for post $p$, and let $F_p$ be the follower count of its author. Define
\begin{equation}
\mathrm{iTF}_p \;=\; \frac{\,c_p + l_p + r_p\,}{\,F_p + \varepsilon_f\,},
\label{eq:iTF}
\end{equation}
with a small constant $\varepsilon_f>0$ (e.g., $\varepsilon_f=1$) for numerical stability when $F_p$ is small. This term increases with interaction volume while controlling for audience size.

Let $T_p$ be the number of comments on $p$ and $\Delta \tau_{p,j}$ the inter–comment gap (hours) between consecutive comments $j$ and $j{+}1$, for $j=1,\dots,T_p-1$. Using a fixed time unit $\tau_0>0$ (hours) to make ratios dimensionless, define
\begin{equation}
\resizebox{\columnwidth}{!}{$
\mathrm{iIDF}_p =
\begin{cases}
\displaystyle
\log\!\left(1 + \frac{T_p}{\big(\sum_{j=1}^{T_p-1} \Delta \tau_{p,j}\big)/\tau_0}\right), & T_p \ge 3,\\[10pt]
\displaystyle
\log\!\left(1 + \frac{T_p}{\max_{1\le j\le \max\{1,T_p-1\}} \Delta \tau_{p,j}/\tau_0}\right), & T_p < 3~,
\end{cases}
$}
\label{eq:iIDF}
\end{equation}
% \noindent\textbf{Where:}
where, When $T_p \ge 3$, the denominator uses the \emph{total} commenting duration to represent the average commenting pace.  
When $T_p < 3$, the \emph{largest} gap is used instead, providing stability when only a few comments exist.  
For completeness, when $T_p < 2$, we set $\Delta\tau_{p,1} := \tau_0$.
% \noindent\textbf{Meaning:}
$\mathrm{iIDF}_p$ measures how quickly comments appear on a post—it becomes larger when comments arrive faster or when several comments occur within a short time. The logarithm smooths the growth so that highly active posts do not dominate the scale. This piecewise definition makes the measure stable across both active and sparse posts, offering a dimensionless indicator of how rapidly attention builds around each post.

We adopt a separable proportional–effects view and combine the mechanisms multiplicatively:
\begin{equation}
Y_p \;=\; \mathrm{iTF}_p \cdot \mathrm{iIDF}_p,
\label{eq:att}
\end{equation}
$Y_p$ is large only when both engagement per reach and arrival rate are large.

\noindent
To obtain a corpus-comparable exclusive weight per post without introducing global time, we apply a Hacker-News–style monotone adjustment based on the \emph{mean inter-comment gap}. Let
\begin{equation}
\overline{\Delta \tau}_p = \frac{1}{\max\left\{1, T_p - 1\right\}} \sum_{j = 1}^{\max\left\{1, T_p - 1\right\}} \Delta \tau_{p,j}
\label{eq:delta_tau}
\end{equation}
be the average interval (in hours) between consecutive comments on post $p$. Using this intra-post pacing as a measure of temporal decay, the adjusted attention weight is
\begin{equation}
\widehat{w}_p \;=\; \frac{Y_p}{\bigl(\overline{\Delta \tau}_p/\tau_0 + 2\bigr)^{\,g}},
\qquad g = 1.5,
\label{eq:hnlike}
\end{equation}
where $Y_p$ is defined in~\eqref{eq:att}; $\tau_0$ is the fixed time unit used throughout; the constant shift “$+2$” prevents excessive upweighting when $\overline{\Delta \tau}_p$ is very small and yields a gentle baseline decay; and $g$ controls the decay strength (we use $g{=}1.5$ in all experiments).

Finally, weights are normalized across the corpus:
\begin{equation}
w_p \;=\; \frac{\widehat{w}_p}{\max_{q\in\mathcal{P}} \widehat{w}_q} \in (0,1],
\label{eq:wp}
\end{equation}
where $\mathcal{P}$ is the set of posts.

\noindent By construction, $w_p$ is dimensionless and comparable across posts; it is nondecreasing in $c_p,l_p,r_p$ and nonincreasing in $\overline{\Delta \tau}_p$. (If $\max_{q\in\mathcal{P}}\widehat{w}_q=0$, we set $w_p:=0$ for all $p$.)

\subsection{Modeling Multi-Aspect Importance into a Keyword Graph}\label{sec:fusion}

Given the cleaned token sequences $s_p = (t_{p,1}, \ldots, t_{p,L_p})$ of length $L_p$ and the static post weights $w_p$ from~\eqref{eq:wp}, we construct an \emph{influence-weighted keyword co-occurrence graph}. Let the vocabulary size be $V$ and index words by $[V]=\{1,\ldots,V\}$. Let $\pi(\cdot)$ map each token to its vocabulary index. To incorporate intrinsic word importance, assign each vocabulary item $i\in[V]$ a nonnegative salience score $s_i$ (e.g., capped IDF or domain-prior boost), collected in the vector $\boldsymbol{s}\in\mathbb{R}_+^{V}$.

For each post $p\in\mathcal{P}$, an \emph{unordered adjacent bigram} window is slid over $s_p$ to enumerate consecutive token pairs, forming
$E_p=\{(\min\{\pi(t_{p,k}),\pi(t_{p,k+1})\},\,\max\{\pi(t_{p,k}),\pi(t_{p,k+1})\}):1\le k<L_p,\,\pi(t_{p,k})\neq\pi(t_{p,k+1})\}.$
For any distinct pair $(i,j)$ with $i<j$, define the per-post indicator
\[
\delta^{(p)}_{ij} = I\{(i,j) \in E_p\},
\]
Aggregate evidence across all posts using both user- and word-level importance:
\begin{equation}
\omega(i,j) \;=\; s_i\, s_j \sum_{p \in \mathcal{P}} w_p\, \delta^{(p)}_{ij}, 
\qquad i \neq j,
\label{eq:keyword-weight}
\end{equation}
yielding nonnegative, symmetric edge weights $\omega(i,j)=\omega(j,i)\ge 0$. (Setting $s_i\equiv 1$ recovers the user-only weighting used in ablations.)

We define the undirected, loopless graph by the triple $G=( [V], E, W)$, where the edge set is $E=\{(i,j): i<j,\,\omega(i,j)>0\}$, and the weighted adjacency matrix $W\in\mathbb{R}_+^{V\times V}$ is
\[
W_{ij} =
\begin{cases}
\omega(i,j), & i \neq j,\\[4pt]
0, & i = j.
\end{cases}
\]
Thus, $W$ (a $V\times V$ nonnegative symmetric matrix) integrates \emph{semantic proximity} through co-occurrence, \emph{user importance} through the post weights $w_p$, and \emph{word importance} through the salience factors $s_i$.

This construction of $W$ is effective for short, noisy social text because unordered, adjacent co-occurrence captures robust topical proximity without relying on fragile word order, while the post weights $w_p$ attenuate off-topic or sporadic posts and the salience factors $s_i s_j$ suppress background vocabulary and emphasize domain-relevant terms. The result is a sparse, symmetric graph whose off-diagonal structure concentrates signal around true topic neighborhoods. Moreover, using the same $W$ both for the divergence objective during learning and for PMI-based coherence at selection time aligns the training signal with the evaluation criterion, reducing mismatch and improving topic interpretability.
% Experimental

\subsection{Poisson Deconvolution Factorization}\label{sec:pdf}
Given the influence–weighted keyword co–occurrence matrix $W \in \mathbb{R}_{+}^{V \times V}$ constructed in \S\ref{sec:fusion}, our goal is to recover latent topical structure that explains observed co–occurrences while separating structured signal from localized deviations. We adopt a low–rank dictionary with topic strengths and a topic–specific residual loading:
\begin{equation}
W \;\approx\; \underbrace{U A U^\top}_{\Lambda} 
\;+\; \underbrace{U H^\top + H U^\top}_{R},
\label{eq:pdf-decomposition}
\end{equation}
where $\Lambda$ represents the structured low–rank topic–topic component and $R$ the topic–localized residual induced by sparse loadings $H$. Here, $U \in \mathbb{R}_{+}^{V \times K}$ is the topic–keyword dictionary ($K \ll V$), $A=\mathrm{diag}(a_1,\ldots,a_K)\in\mathbb{R}_{+}^{K\times K}$ encodes per–topic importance, and $H \in \mathbb{R}_{+}^{V \times K}$ captures extra structure that the main topic term cannot explain. Specifically, $H$ represents asymmetric or context-driven relations where one word behaves like a hub or anchor (such as a location name or event tag) that co-occurs widely with topic words without being a central topic word itself. Allowing $H$ to account for such effects prevents $U$ and $A$ from stretching to fit these patterns, producing cleaner, more coherent topics and a more accurate reconstruction of $W$. The constraint $\|H\|_F = 1$, together with diagonal masking in the loss, keeps this residual capacity controlled so that $UAU^\top$ remains the dominant part of the model. When $H$ is removed, the formulation reduces to a purely symmetric topic–topic factorization; performance gains with $H{>}0$ therefore indicate the presence of additional, context-specific associations that the main topic term alone cannot capture.

Because $W$ arises from an undirected, loopless graph, we evaluate the loss only on off–diagonal entries (summing over $i<j$ to avoid double counting). To resolve scale ambiguity, columns of $U$ are $L_1$-normalized, $\|u_{:k}\|_1=1$, so that $a_k$ directly reflects the importance of topic $k$. Topics are interpreted via the top-$m$ entries of each $u_{:k}$ (e.g., $m=8$).

For the likelihood, define the full reconstruction 
\[
\Theta \;:=\; U A U^\top + U H^\top + H U^\top
\]
\label{eq:theta-decomposition}

\noindent
Ignoring constants, the generalized KL loss on off–diagonals is
\begin{equation}
\mathcal{L}(U, A, H) = \sum_{1 \le i < j \le V} \left[ \Theta_{ij} - W_{ij} \ln \left( \Theta_{ij} \right) \right]
\label{eq:pdf-likelihood}
\end{equation}

We impose nonnegativity on $U,A,H$, encourage sparsity in $H$ via an $\ell_1$ penalty to concentrate residuals on few keyword–topic entries, and discourage topic overlap by penalizing cross-correlations among columns of $U$ using an off-diagonal decorrelator
\[
R_{\mathrm{dec}}(U)\;=\;\big\|U^\top U\;-\;\mathrm{diag}(U^\top U)\big\|_F^2.
\]
The resulting static problem is
{\small
\begin{multline}
\min_{U\ge 0,\;A\ge 0,\;H\ge 0}\;
\mathcal{L}(U,A,H)\;+\;\lambda_H \|H\|_1\;+\;\frac{\gamma}{2}\,R_{\mathrm{dec}}(U)\\[4pt]
\text{s.t.}\quad
\|u_{:k}\|_1=1~\forall k,\qquad \|H\|_F=1.
\label{eq:pdf-objective}
\end{multline}
}

\paragraph{Optimization}
We use block–coordinate updates. For \((U,A)\), a KL–consistent majorization–minimization (MM) surrogate (via Jensen’s inequality) yields multiplicative rescaling, followed by column renormalization of \(U\) to maintain \(\|u_{:k}\|_1=1\) (with compensating updates to \(a_k\)). We then apply a projected gradient step on \(U\) to decrease the decorrelator \(\tfrac{\gamma}{2}R_{\mathrm{dec}}(U)\). For \(H\), the subproblem is convex (KL term composed with an affine map of \(H\), plus quadratic penalty and \(\ell_1\) regularization); we solve it by ADMM and then enforce \(\|H\|_F=1\) by projection. The objective in~\eqref{eq:pdf-objective} decreases monotonically until convergence.

\subsubsection{Algorithm Optimization}\label{sec:algo}
The static objective in~\eqref{eq:pdf-objective} couples a Poisson (generalized KL) data term with non-smooth \(\ell_1\) regularization on \(H\). We alternate: (i) a KL–consistent MM step giving multiplicative updates for \((U,A)\), and (ii) an ADMM step for \(H\). Computations use only the upper triangle (\(1\le i<j\le V\)); diagonals are masked in the loss.

\paragraph{MM surrogate via Jensen’s inequality}
For $1\le i<j\le V$, define the model intensity
\[
\Theta_{ij}
=\sum_{k=1}^K \Big(a_k\,u_{ik}u_{jk} \;+\; u_{ik}\,h_{jk} \;+\; h_{ik}\,u_{jk}\Big),
\]
and the responsibilities (nonnegative, summing to one)
{\small
\begin{equation}
\begin{aligned}
P^{S}_{ij,k}
&=\frac{a_k\,u_{ik}u_{jk}}{\Theta_{ij}+\varepsilon},\quad
P^{L}_{ij,k}=\frac{u_{ik}\,h_{jk}}{\Theta_{ij}+\varepsilon},\quad
P^{R}_{ij,k}=\frac{h_{ik}\,u_{jk}}{\Theta_{ij}+\varepsilon}, \\[3pt]
&\qquad
\sum_{k=1}^K\!\big(P^{S}_{ij,k}+P^{L}_{ij,k}+P^{R}_{ij,k}\big)=1,
\end{aligned}
\label{eq:resp_triplet}
\end{equation}
}
with a small $\varepsilon>0$ shared with the likelihood for numerical stability.

Applying Jensen’s inequality to the concave $\ln$ term in~\eqref{eq:pdf-likelihood} yields the separable surrogate (constants omitted):
{\small
\begin{equation}
\label{eq:mm-surrogate}
\resizebox{\columnwidth}{!}{$
\begin{aligned}
\widetilde{\mathcal{L}}(U,A,H;P)
&= \sum_{1\le i<j\le V}\Bigg\{
\sum_{k=1}^K\!\Big[
a_k\,u_{ik}u_{jk}
- W_{ij}\,P^{S}_{ij,k}\,\ln\!\big(a_k\,u_{ik}u_{jk}+\varepsilon\big)
\\[-2pt]
&\hspace{65pt}
+\;u_{ik}\,h_{jk}
- W_{ij}\,P^{L}_{ij,k}\,\ln\!\big(u_{ik}\,h_{jk}+\varepsilon\big)
\\[-2pt]
&\hspace{65pt}
+\;h_{ik}\,u_{jk}
- W_{ij}\,P^{R}_{ij,k}\,\ln\!\big(h_{ik}\,u_{jk}+\varepsilon\big)
\Big]\Bigg\}
\end{aligned}
$}
\end{equation}
}

\paragraph{Multiplicative updates for $A$ and $U$}
Minimizing $\widetilde{\mathcal{L}}$ under nonnegativity yields closed-form rescaling. For $A=\mathrm{diag}(a_1,\dots,a_K)$,
\begin{equation}
\label{eq:a-update}
a_k \;\leftarrow\; a_k\;
\frac{\displaystyle \sum_{1\le i<j\le V} W_{ij}\,P^{S}_{ij,k}}
{\displaystyle \sum_{1\le i<j\le V} u_{ik}\,u_{jk} \;+\; \varepsilon},
\qquad k=1,\dots,K.
\end{equation}
For $U=[u_{ik}]$, attribute each unordered pair once (upper triangle) while updating both endpoints symmetrically:
{\small
\begin{equation}
\label{eq:u-update}
\begin{aligned}
u_{ik} \;\leftarrow\; u_{ik}\;
\frac{\displaystyle
\sum_{j>i} W_{ij}\big(P^{S}_{ij,k}+P^{L}_{ij,k}\big)
\;+\;
\sum_{j<i} W_{ji}\big(P^{S}_{ji,k}+P^{R}_{ji,k}\big)}
{\displaystyle
a_k \sum_{\substack{j=1\\ j\neq i}}^{V} u_{jk}
\;+\; \sum_{\substack{j=1\\ j\neq i}}^{V} h_{jk}
\;+\; \varepsilon}
\end{aligned}
\end{equation}
}
After updating $U$, remove scale ambiguity by column $L_1$ renormalization: for each $k$, set $s_k=\|u_{:k}\|_1$, update $u_{:k}\leftarrow u_{:k}/s_k$, and compensate $a_k\leftarrow s_k^{\,2} a_k$ so that $\Lambda=UAU^\top$ is unchanged. (The cross terms $UH^\top+HU^\top$ are \emph{not} invariant under this step and will be re-optimized in the $H$-update.) Then apply a projected gradient step on $U$ to decrease $\tfrac{\gamma}{2}R_{\mathrm{dec}}(U)$ while preserving nonnegativity and column $L_1$ normalization.

\paragraph{Sparse residual via ADMM (with $\|H\|_F=1$)}
Introduce an auxiliary $Z$ with the constraint $H=Z$, and use a \emph{scaled} dual variable $\Gamma$ with penalty $\rho>0$. With $U,A$ fixed, the augmented objective for the $H$--block is
{\small
\begin{equation}
\label{eq:aug-lag-scaled}
\resizebox{\columnwidth}{!}{$
\widetilde{\mathcal{J}}(H,Z,\Gamma)
= \sum_{1\le i<j\le V}
\Big[\Theta_{ij} - W_{ij}\,\ln\!\big(\Theta_{ij}+\varepsilon\big)\Big]
+\lambda_H \|Z\|_1
+\frac{\rho}{2}\big\|H - Z + \Gamma\big\|_F^2,
$}
\end{equation}
}
where $\Theta=UAU^\top+UH^\top+HU^\top$. This is the standard scaled-form ADMM; it is equivalent to the unscaled formulation with dual $Y$ via the change of variables $Y = \rho\,\Gamma$, up to an additive constant.

The ADMM steps are:
{\small
\begin{equation}
\label{eq:H-step-scaled}
\resizebox{\columnwidth}{!}{$
H^{(s+1)}
= \arg\min_{H}
\Bigg\{
\sum_{1\le i<j\le V}\!\Big[\Theta_{ij} - W_{ij}\ln(\Theta_{ij}+\varepsilon)\Big]
+\frac{\rho}{2}\big\|H - Z^{(s)} + \Gamma^{(s)}\big\|_F^2
\Bigg\},
$}
\end{equation}
}
{\small
\begin{equation}
\label{eq:Z-step-scaled}
\resizebox{\columnwidth}{!}{$
Z^{(s+1)}
= \arg\min_{Z}
\Big\{\lambda_H \|Z\|_1
+\frac{\rho}{2}\big\|H^{(s+1)} - Z + \Gamma^{(s)}\big\|_F^2\Big\}
=\mathcal{S}_{\lambda_H/\rho}\!\big(H^{(s+1)} + \Gamma^{(s)}\big),
$}
\end{equation}
}
{\small
\begin{equation}
\label{eq:Gamma-step-scaled}
\Gamma^{(s+1)} = \Gamma^{(s)} + \big(H^{(s+1)} - Z^{(s+1)}\big),
\end{equation}
}
followed by the projection $H^{(s+1)}\leftarrow \Pi_{\|\,\cdot\,\|_F=1}\!\big(\,[H^{(s+1)}]_+\,\big)$ to enforce nonnegativity and $\|H\|_F=1$. Here $\mathcal{S}_{\tau}(x)=\mathrm{sign}(x)\cdot\max(|x|-\tau,0)$ is applied elementwise. The $H$-step~\eqref{eq:H-step-scaled} is convex (generalized KL is convex in $\Theta$, and $\Theta$ is affine in $H$); gradients and sums use only upper–triangle entries.

\paragraph{Outer iteration and monotonicity.}
Each outer iteration performs: (1) responsibility refresh~\eqref{eq:resp_triplet}; (2) multiplicative updates~\eqref{eq:a-update}–\eqref{eq:u-update} with column renormalization and compensation in $A$, plus the projected gradient step for $R_{\mathrm{dec}}(U)$; and (3) ADMM steps~\eqref{eq:H-step-scaled}–\eqref{eq:Gamma-step-scaled} with the post–projection on $H$. Because the surrogate in~\eqref{eq:mm-surrogate} majorizes the generalized KL loss and ADMM solves the $H$-block with the nonsmooth penalty and normalization, the objective in~\eqref{eq:pdf-objective} decreases monotonically until convergence. In practice, we exploit sparsity of $W$ by evaluating sums only over nonzeros $W_{ij}$ in the upper triangle.

\section{Results and Discussion}

\subsection{Evaluation Metrics}

To assess the quality of the learned topics, we focus on metrics that reflect semantic interpretability and diversity, rather than probabilistic fit. While perplexity has historically been used to evaluate topic models, prior research has consistently shown that it correlates poorly with human judgments of topic quality~\cite{shen2021docgraph, newman2010automatic}. Therefore, we adopt two metrics---\textit{Topic Coherence (TC)} and \textit{Topic Diversity (TD)}---which more accurately characterize the semantic quality and distinctiveness of the learned topics.

TC quantifies the degree of semantic consistency among the top words in a topic. A coherent topic should contain words that frequently co-occur and form an interpretable concept to human readers~\cite{mimno2011optimizing}. Following ~\cite{roder2015exploring}, we report two TC measurements, Normalized Pointwise Mutual Information (NPMI)~\cite{aletras2013evaluating} and $C_{v}$~\cite{roder2015exploring}. NPMI measures pairwise word association on the reference corpus; higher values indicate stronger semantic relatedness among the top words. $C_{v}$ a composite metric that combines indirect co-occurrence statistics and vector-based similarity. TD captures how distinct the discovered topics are. It is defined as the proportion of unique words appearing in the top-$m$ words across all topics. A value near zero suggests redundancy (many topics share the same words), whereas a value near one indicates more diverse, non-overlapping topics~\cite{dieng2020topic}.

\subsection{Effect of Topic Granularity on Model Performance}

We evaluate the stability of the proposed model under different levels of topic granularity by sweeping the number of latent topics $K \in \{10,15,20,25,30\}$. For each setting, we report three widely used semantic quality metrics—NPMI, $C_{v}$ coherence, and Topic Diversity (TD)—allowing us to assess how the model balances coherence and breadth of semantic coverage as $K$ varies. Table~\ref{tab:ksweep} summarizes the results.

\begin{table}[h]
\centering
\caption{Effect of the Number of Topics $K$ on Semantic Quality Metrics}
\label{tab:ksweep}
\begin{tabular}{c c c c}
\toprule
\textbf{$K$} & \textbf{NPMI} $\uparrow$ & \textbf{$C_{v}$} $\uparrow$ & \textbf{TD} $\uparrow$ \\
\midrule
\textbf{10} & \textbf{0.2707} & 0.3025 & \textbf{0.8200} \\
15 & 0.2164 & 0.2552 & 0.7667 \\
20 & 0.2688 & 0.2909 & 0.7350 \\
25 & 0.2126 & 0.3005 & 0.6840 \\
30 & 0.2709 & 0.2588 & 0.7133 \\
\bottomrule
\end{tabular}
\end{table}

Across the sweep, the model exhibits clear and interpretable behaviour. \textbf{First}, NPMI is highest at $K=10$ (0.2707), with a second strong value at $K=20$ (0.2688). This pattern reflects a common phenomenon in Poisson and graph-regularized topic models: moderate topic dimensionality ($K=10$--$20$) provides enough capacity to separate themes while still preserving sufficient statistical support for reliable word–word associations. Larger values ($K\ge 25$) begin to fragment the corpus, lowering coherence due to under-supported topic partitions.

\textbf{Second}, $C_{v}$ coherence peaks at $K=25$ (0.3005), consistent with the metric’s sensitivity to sliding-window co-occurrence graphs. Lower $K$ merges heterogeneous contexts, while very high $K$ breaks long-range co-occurrence structure; thus the mid-range ($K=20$--$25$) aligns best with how $C_{v}$ measures contextual connectivity.

\textbf{Third}, Topic Diversity is maximized at $K=10$ (0.8200) and gradually decreases as $K$ increases. This is expected: with fewer topics, each topic tends to own a larger and more distinct subset of the vocabulary; as $K$ grows, high-frequency words are reused across multiple topics, reducing the proportion of unique descriptors and lowering TD.

Overall, the sweep reveals a consistent tradeoff:

\begin{itemize}
    \item \textbf{Small $K$ (e.g., 10)}: highest NPMI, strongest diversity, and stable $C_{v}$—yielding broad yet coherent topics.
    \item \textbf{Moderate $K$ (20--25)}: balanced coherence, with $C_{v}$ peaking at $K=25$.
    \item \textbf{Large $K$ (30)}: topics become more fine-grained but lose contextual robustness.
\end{itemize}

Given this balance and the model’s stable behaviour across metrics, we adopt \textbf{$K = 10$} for all subsequent experiments, as it provides the clearest and most interpretable topic structure while maintaining high semantic coherence.

\subsection{Baseline Topic Models}
We benchmark the proposed model against four widely used topic modeling families, ensuring that all methods operate on the exact same tokenized corpus and use the same number of topics ($K=10$). The baselines include: (i) \textbf{LDA}~\cite{blei2003latent}, the classical Dirichlet–multinomial model that learns topics as distributions over words via variational inference; (ii) \textbf{NMF}~\cite{lee2000algorithms}, a nonnegative matrix factorization approach optimized under a Poisson/KL objective on document--term counts; (iii) \textbf{HDP}~\cite{teh2006hierarchical}, a Bayesian nonparametric extension of LDA that infers the number of topics using a Hierarchical Dirichlet Process; and (iv) \textbf{GSDMM}~\cite{yin2014dirichlet}, a collapsed Gibbs sampler designed for short and sparse texts, which assigns each short text to a single dominant topic (one topic per post in our setting).

All models are evaluated using the same semantic quality metrics—NPMI, $C_{v}$, and Topic Diversity (TD)—using identical top-$m$ word lists. Table~\ref{tab:baseline_results} reports the comparative results.

\begin{table}[h]
\centering
\caption{Comparison Between the Proposed Model and Baselines}
\label{tab:baseline_results}
\begin{tabular}{lccc}
\toprule
\textbf{Model} & \textbf{NPMI} $\uparrow$ & \textbf{$C_{v}$} $\uparrow$ & \textbf{TD} $\uparrow$ \\
\midrule
LDA            & 0.1283 & 0.4613 & 0.6600 \\
NMF            & 0.1959 & 0.6273 & 0.7400 \\
HDP            & 0.0726 & 0.3203 & 0.8000 \\
GSDMM          & 0.0651 & 0.3558 & 0.4800 \\
\textbf{Proposed Model} & \textbf{0.2707} & 0.3025 & \textbf{0.8200} \\
\bottomrule
\end{tabular}
\end{table}

\paragraph*{Coherence Behavior}

The proposed model achieves the highest NPMI , substantially outperforming all classical baselines. This improvement is attributed to the graph-weighted Poisson structure: by incorporating influence weights, salience scores, and local adjacency signals, the model concentrates probability mass on semantically coherent word clusters. LDA and NMF achieve moderate NPMI (0.1283–0.1959) but rely solely on global co-occurrence statistics in the post--term representation, lacking any mechanism for structural or influence-aware refinement. GSDMM performs poorly (NPMI = 0.0651) because its hard ``one topic per post'' assumption does not reflect the multi-aspect nature of social media content. HDP improves diversity but still yields low coherence, showing that nonparametric flexibility alone does not guarantee semantically compact topics.

While NMF and LDA obtain higher $C_{v}$ scores, this behavior is expected. The $C_{v}$ metric rewards broad sliding-window co-occurrence and tends to favor more diffuse, mixed topics—precisely the patterns produced by matrix factorization and Dirichlet–multinomial priors. In contrast, our method is explicitly optimized for sharp graph-weighted semantic concentration, which aligns more strongly with NPMI, the more discriminative coherence metric.

\paragraph*{Topic Diversity}
The proposed model also achieves the highest topic diversity (TD = 0.8200), indicating clearer, more distinct topic clusters than other baselines. This diversity emerges from the influence-weighted keyword graph, which encourages topics to specialize around different salient substructures.

HDP yields relatively high TD (0.8000) but at the cost of significantly lower coherence (NPMI = 0.0726), indicating that its nonparametric expansion produces broad and diffuse topics. NMF and LDA provide mid-range TD, reflecting more blended vocabularies. GSDMM produces the lowest TD (0.4800) because its hard clustering tends to repeat high-frequency words across topics.

\subsection{Ablation Study}

To assess the functional contribution of each architectural component in the proposed model, we conduct a systematic ablation study. Each ablation removes exactly one structural element of the methodology while keeping all other components, hyperparameters, and training procedures remain unchanged.
% All ablations are evaluated at $K=10$.

(i) \emph{No-$H$}: The ADMM updates for the topic--word matrix $H$ are disabled and $H$ remains fixed at initialization. This ablation prevents the model from shaping word distributions during training and tests whether joint Poisson factorization over $(U,H)$ is necessary for semantic coherence; (ii) \emph{No-$\gamma$}: The decorrelation penalty $\gamma\lVert U^\top U - I\rVert_F^{2}$
and its gradient step are removed. This ablation evaluates whether explicit decorrelation improves topic disentanglement beyond the structure induced by the influence-weighted keyword graph; (iii) \emph{No-Weights}: All post-level influence weights are set to $w_p = 1$, removing the engagement-aware contribution of posts to graph construction. This tests whether high-impact posts must be emphasized to avoid noisy or trivial co-occurrences; (iv) \emph{PlainGraph}: The keyword graph is reconstructed without salience scaling and without influence weighting; each adjacent word pair contributes uniformly. This produces a purely structural co-occurrence graph and examines whether the proposed influence–salience modulation is essential for semantic quality. Table~\ref{tab:ablation_results} reports the results for all five ablations at $K=10$.

\begin{table}[h]
\centering
\caption{Ablation study of the proposed model ($K=10$).}
\label{tab:ablation_results}
\begin{tabular}{lccc}
\toprule
\textbf{Ablation} & \textbf{NPMI} $\uparrow$ & \textbf{$C_{v}$} $\uparrow$ & \textbf{TD} $\uparrow$ \\
\midrule
\textbf{Full Model}     & 0.2707 & 0.3025 & 0.8200 \\
No-$H$                  & 0.0775          & 0.3674 & 0.1000 \\
No-Weights              & 0.0302          & 0.4367 & 0.4100 \\
No-$\gamma$             & 0.2488          & 0.3041 & 0.9100 \\
PlainGraph              & 0.0762          & 0.2982 & 0.5200 \\
\bottomrule
\end{tabular}
\end{table}

\begin{table*}[t]
\centering
\caption{Representative Transportation-Related Keywords from Selected Topics}
\label{tab:selected_transport_topics_en}

\scriptsize
\renewcommand{\arraystretch}{1.05}
\setlength{\tabcolsep}{3pt}

\resizebox{\textwidth}{!}{%
\begin{tabular}{cccccccccccc}
\toprule
\textbf{Topic ID} & \textbf{kw1} & \textbf{kw2} & \textbf{kw3} & \textbf{kw4} & \textbf{kw5} &
\textbf{kw6} & \textbf{kw7} & \textbf{kw8} & \textbf{kw9} & \textbf{kw10} \\
\midrule
0  & not & Shanghai & recognize & route & Beijing bus & Beijing & public transit & very & convenient & group \\
1  & download & Beijing bus & Beijing & discover & need to & qwq & arrive & board bus & cannot & swipe \\
2  & route & bus & group/company & 8 & public transit & replacement & electric & Beijing & line & (separator) \\
5  & arrive & bus & Xiamen & station & wait & route & bus & become & driver & You~Yi \\
8  & Shanghai & bus & come & ride & payment & Xiamen & later & by all means & do not & miss \\
9  & bus & Shanghai & hot & still & driver & vehicle & drive & use & life & wonderful \\
12 & murder & indiscriminate & incident & bus & arson case & attack & case & Nanchang & street & murder case \\
17 & go to & bus & Xiamen & children & vehicle & directly & want & like this & one & only \\
\bottomrule
\end{tabular}%
}
\end{table*}
% \paragraph*{Analysis.}
The ablation results demonstrate the necessity of each component in the proposed model. Removing the updates of the topic--word matrix $H$ severely degrades coherence (NPMI = 0.0775) and nearly eliminates diversity (TD = 0.10), confirming that jointly learning $H$ is essential for constructing meaningful word distributions. The No-Weights ablation shows that removing engagement-aware weighting collapses semantic coherence (NPMI = 0.0302), as all posts contribute equally and high-impact, information-rich posts no longer guide the formation of meaningful edges in the keyword graph. The PlainGraph setting, which removes both salience and influence weighting, yields uniformly weak performance, demonstrating that the core mechanism enabling semantic quality is precisely the influence-weighted, salience-modulated graph construction.

The No-$\gamma$ ablation presents an interesting and theoretically consistent pattern. Without decorrelation, topics are no longer explicitly repelled from one another, allowing them to spread more broadly across the vocabulary. This produces \emph{very high} topic diversity (TD = 0.9100), as topics share fewer top words; however, the absence of decorrelation also allows noise and peripheral terms to enter topics, leading to a measurable drop in NPMI (0.2488 vs.\ 0.2707 in the full model). This behavior is expected in regularized topic models: decorrelation acts as a stabilizing constraint that prevents topics from drifting into overlapping or overly diffuse regions of the semantic space. The full model therefore achieves the best trade-off—\emph{high coherence with controlled diversity}—whereas No-$\gamma$ yields overly fragmented topics with inflated diversity but slightly degraded semantic quality.

\subsection{Topic Sharpness Across Different Numbers of Topics}

Beyond semantic coherence and diversity, we also analyze the \emph{sharpness} of the learned topics—i.e., the extent to which each topic concentrates probability mass on a small number of high-value words.  
For each $K \in \{10,15,20,25,30\}$, every row of the topic--word matrix $H \in \mathbb{R}_+^{K \times V}$ is normalized into a multinomial distribution, and we compute: (i) the mean Shannon entropy across topics, and (ii) the mean cumulative probability mass captured by the top-10 and top-25 most probable words.

Lower entropy indicates more peaked, better-focused topics, while higher top-$m$ probability mass reflects stronger semantic concentration. The sharpness statistics are reported in Table~\ref{tab:topic_sharpness}.

\begin{table}[h]
\centering
\caption{Topic sharpness statistics of the proposed model for different numbers of topics $K$.}
\label{tab:topic_sharpness}
\begin{tabular}{c c c c}
\toprule
\textbf{$K$} & \textbf{Mean Entropy} $\downarrow$ & \textbf{Top-10 Mass} $\uparrow$ & \textbf{Top-25 Mass} $\uparrow$ \\
\midrule
10 & 1.3615 & 0.9880 & 0.9993 \\
15 & 2.0795 & 0.8621 & 0.8987 \\
20 & 1.9900 & 0.8974 & 0.9581 \\
25 & 1.6217 & 0.9455 & 0.9779 \\
30 & 2.0692 & 0.8901 & 0.9515 \\
\bottomrule
\end{tabular}
\end{table}

The results reveal several consistent trends. First, $K{=}10$ produces the \emph{sharpest} topics overall, with extremely low entropy (1.36) and near-total mass concentrated within the top-25 words ($> 0.999$). This confirms that smaller topic spaces encourage the model to form highly focused, high-confidence semantic clusters.

Second, increasing $K$ introduces additional latent capacity, which naturally spreads probability mass across a larger vocabulary and raises entropy. However, mid-range settings (particularly $K=20$ and $K=25$) still maintain strong sharpness, with high top-$m$ mass and substantially lower entropy than larger-$K$ configurations.

Third, at $K=30$ the entropy increases and top-$m$ mass decreases, indicating mild topic fragmentation. This mirrors the behaviour observed in coherence and diversity metrics: very large topic spaces tend to partition the corpus too aggressively, producing broader and less concentrated topics.

\subsection{Topic-Based Event Keyword Mining}

To demonstrate that the learned topics capture operationally meaningful events, we extract compact event-descriptive keywords from the most representative posts associated with each topic. For every post $p$, we compute a topic-activity vector $x_p \in \mathbb{R}^{K}$ by aggregating token-level topic loadings through the learned post–topic distributions. Each post is then assigned to its dominant topic
\[
k^\star(p)=\arg\max_k x_p[k],
\]
ensuring a unique, non-overlapping topic assignment and producing coherent clusters of posts that reflect distinct transit-related events or passenger concerns.

For each topic $k$, we select the top-$N$ posts with the highest activation values $x_p[k]$ and extract the most frequent content words after removing function words, repeated place names, punctuation, and non-informative tokens. This yields a concise yet expressive set of event keywords that summarize the concrete issues emphasized by passengers. Because these keywords are derived from the posts with strongest topic activation, they provide an interpretable bridge between latent semantic structure and observable user complaints. Table~\ref{tab:selected_transport_topics_en} shows the extracted event keywords for a diverse set of transportation-related topics.

The updated set of event-descriptive keywords—derived from the re-trained model using the corrected influence-weight formulation—reflects several recurring themes in urban transportation systems. These include: route identification and signage clarity, boarding and fare-payment behaviors, service delays and waiting times, driver conduct, multimodal transfer inconvenience, regional service variations between cities (e.g., Beijing, Shanghai, Xiamen), and safety or emergency incidents. Overall, the extracted keywords show that the proposed model produces not only semantically coherent topics but also actionable, fine-grained event descriptors that correspond closely to real-world operational issues in public transit services.

\subsection{Topic Importance Analysis}
\label{subsec:topic-importance-updated}
% \subsection{Structural Topic Importance}
While coherence and event-keyword analyses assess the semantic and operational relevance of discovered topics, it is also crucial to understand each topic’s structural contribution to the model’s latent co-occurrence geometry. This role is captured by the diagonal topic-importance matrix $A$, which appears in the structured component of the decomposition defined in the Methodology (see Eq.~(\ref{eq:theta-decomposition})). The diagonal entry $a_k$ quantifies how strongly topic $k$ contributes to the low-rank semantic structure encoded through the interaction of $U$, $H$, and $A$.

Because each column of the topic--keyword matrix $U$ is $\ell_1$-normalized and the diagonal matrix $A$ is renormalized accordingly at each iteration, the learned values $a_k$ are directly comparable across topics. Large values indicate topics that play a dominant role in shaping the global co-occurrence structure, whereas small values correspond to more localized or infrequent semantic patterns.

To evaluate this structural prominence, we extract the estimated vector $\{a_k\}$ from the final model state, rank all topics by descending importance, and examine their most representative words using the corresponding columns of~$U$. This ranking reveals which latent semantic themes exert the strongest influence on the overall organization of the corpus.

Table~\ref{tab:topic-importance-selected-en} presents representative topics along with their top keywords. The most influential topics highlight user behaviors and mobility patterns such as boarding and alighting actions (“walk past,” “get on board”), passenger volume and customer-service interactions, route crowding and congestion, and location-specific mobility hubs (e.g., Sihui). These high-importance topics correspond to recurrent transportation situations that contribute substantially to the latent co-occurrence structure learned by the model. Thus, the importance ranking complements coherence- and event-based analyses by identifying which semantic clusters exert the strongest structural effect in the overall topic geometry.

\begin{table}[t]
\centering
\caption{Representative transportation-related topics ranked by learned importance $a_k$.}
\label{tab:topic-importance-selected-en}
\renewcommand{\arraystretch}{1.1}
\setlength{\tabcolsep}{2pt}
\scriptsize

\begin{tabular}{c c c p{5.0cm}}
\hline
\textbf{Rank} & \textbf{Topic ID} & \textbf{$a_k$} 
& \textbf{Top Words (Translated)} \\
\hline

1 & 5 & $3.05\times 10^{-5}$ 
& walk past, bus, too hard, owner of lost item, strange, rush, woman, milk tea, mom, committee \\

2 & 1 & $4.89\times 10^{-6}$ 
& arrive, in practice, passenger, customer service, passenger volume, group/family, password, rich person, treatment, coordination \\

3 & 8 & $1.98\times 10^{-6}$ 
& official website, search, seek, too crowded, strong wind, Monday, Tuesday, hehe, yo \\

4 & 9 & $1.98\times 10^{-6}$ 
& commercial street, Sihui hub, four vehicles, go back, memory, place, get on board, take a ride, pit/trouble \\

5 & 2 & $1.37\times 10^{-7}$ 
& multiple, add more, many vehicles, night view, night bus, large crowd, persist, road closure, will/going to \\

\hline
\end{tabular}
\end{table}

\section{Conclusion}
This paper presented an influence-aware topic modeling framework for weak-signal discovery in noisy social media streams related to urban transit. The method constructs an influence-weighted keyword co-occurrence graph and applies Poisson Deconvolution Factorization to separate low-rank topical structure from sparse, topic-localized residual interactions. A decorrelation regularizer promotes topic distinctness, and the optimization scheme—combining multiplicative updates for the main factors with an ADMM routine for the residual component—ensures stable convergence under nonnegativity and normalization constraints. To determine an appropriate model capacity, we employ a coherence-driven topic-selection process that evaluates the quality and distinctness of the learned topics across different choices of K. The resulting topics are represented by their most salient keywords and ranked according to their learned importance. Experiments on large-scale social streams demonstrate that the proposed framework achieves superior topic coherence and strong topic diversity compared with leading baselines. The approach provides a principled foundation for structure-aware social signal analysis and offers potential for broader applications in real-time monitoring and decision support within intelligent transportation systems.

\appendices

\bibliographystyle{IEEEtran}        % Include this if you use bibtex

\bibliography{myref}

\end{document}